\documentclass[sigconf]{acmart}

\usepackage{soul}
\usepackage{multirow}
\usepackage{booktabs}
\usepackage{graphicx}
\usepackage{subfigure} 
\usepackage{float}
\usepackage{xspace}
\usepackage{bm}
\usepackage{bbding}

\newcommand*{\etc}{\emph{etc}\@\xspace}
\newcommand*{\ie}{\emph{i.e.}\@\xspace}

\AtBeginDocument{%
  \providecommand\BibTeX{{%
    \normalfont B\kern-0.5em{\scshape i\kern-0.25em b}\kern-0.8em\TeX}}}

\setcopyright{acmcopyright}

\copyrightyear{2021}
\acmYear{2021}
\setcopyright{acmcopyright}\acmConference[MM '21]{Proceedings of the 29th ACM
International Conference on Multimedia}{October 20--24, 2021}{Virtual Event, China}
\acmBooktitle{Proceedings of the 29th ACM International Conference on Multimedia
(MM '21), October 20--24, 2021, Virtual Event, China}
\acmPrice{15.00}
\acmDOI{10.1145/3474085.3479202}
\acmISBN{978-1-4503-8651-7/21/10}

\begin{document}
\fancyhead{}

\title{A Multimodal Framework for Video Ads Understanding  }

\author{Zejia Weng, Lingchen Meng, Rui Wang, Zuxuan Wu, Yu-Gang Jiang}

\affiliation{%
Shanghai Key Lab of Intelligent Information Processing, \\ School of Computer Science, Fudan University, Shanghai
  \country{China}
}
\email{{zjweng20, lcmeng20, ruiwang16, zxwu, ygj}@fudan.edu.cn}

\renewcommand{\shortauthors}{Zejia Weng,Lingchen Meng, Rui Wang et al.}

\begin{abstract}
There is a growing trend in placing video advertisements on social platforms for online marketing, which demands automatic approaches to understand the contents of advertisements effectively. Taking the 2021 TAAC competition as an opportunity, we developed a multimodal system to improve the ability of structured analysis of advertising video content. In our framework, we break down the video structuring analysis problem into two tasks, \ie, scene segmentation and multi-modal tagging. In scene segmentation, we build upon a temporal convolution module for temporal modeling to predict whether adjacent frames belong to the same scene. In multi-modal tagging, we first compute clip-level visual features by aggregating frame-level features with NeXt-SoftDBoF. The visual features are further complemented with textual features that are derived using a global-local attention mechanism to extract useful information from OCR (Optical Character Recognition) and  ASR (Audio Speech Recognition) outputs. Our solution achieved a score of 0.2470 measured in consideration of localization and prediction accuracy, ranking fourth in the 2021 TAAC final leaderboard.
  
\end{abstract}

\ccsdesc[500]{Computing methodologies~Scene understanding}
\ccsdesc[500]{Information systems~Multimedia and multimodal retrieval}

\keywords{Scene Segmentation, Multi-Modal Tagging, Global-Local Attention, Temporal Modeling}

\maketitle

\section{Introduction}
Online video advertising is an effective marketing method due to its advantages such as strong flexibility, widespread, low cost, and strong interaction. Therefore, different companies have invested more and more effort to produce online video advertisements, and use different social platforms to accurately deliver them to users.
Due to the important value of video advertising, there are also many studies related to video advertising, including advertisement recommendation, quality monitoring, interruption time estimation, etc. With the rapid development of the 5G field, the number of video advertisements has also increased rapidly, and thus it is critical to understand the structures of video ads automatically and effectively.

In contrast to traditional video ads classification, video ads structure understanding requires the model to be able to segment the advertisements into different scenes correctly, and then perform multi-label classification for each scene. Therefore, in essence, video ads content structuring can be divided into two relatively independent tasks, which are scene segmentation and multi-label classification.

How to effectively identify scene boundaries remains a challenging problem for scene segmentation.  One method is to divide the task into two stages, shot detection and shot merging, as it is beneficial to reduce the number of potential turning points to be judged~\cite{rao2020local} . However, the boundary error caused by shot segmentation will affect the scene segmentation quality, and no shot boundary information is given on the 2021 TAAC data set, making it difficult to develop a convincing shot segmentation method. Therefore, we use an end-to-end scene segmentation approach, directly predicting whether each time position is a scene turning point by observing its surrounding video frames.

Videos are multi-modal in nature, which has motivated extensive work to leverage different modalities like visual and audio for better video understanding~\cite{MVA:audiovisual,wu2016multi,gao2020listen,weng2021hms}. 
Compared to visual and audio information, textual clues are less explored for video understanding. However, in video ads structuring understanding, textual information plays an important role since the dialogue information and subtitles in the ads are oftentimes directly related to content topics, such as product categories, names of the promotional items, types of advertisements, and \etc. To this end, in the 2021 TAAC Challenge, we focus on OCR and ASR features in addition to visual modalities, and propose global-local attention to combine OCR and ASR features to fully exploit textual information for improving the overall performance.

\begin{figure*}[h]
\centering
\includegraphics[width=0.9\linewidth]{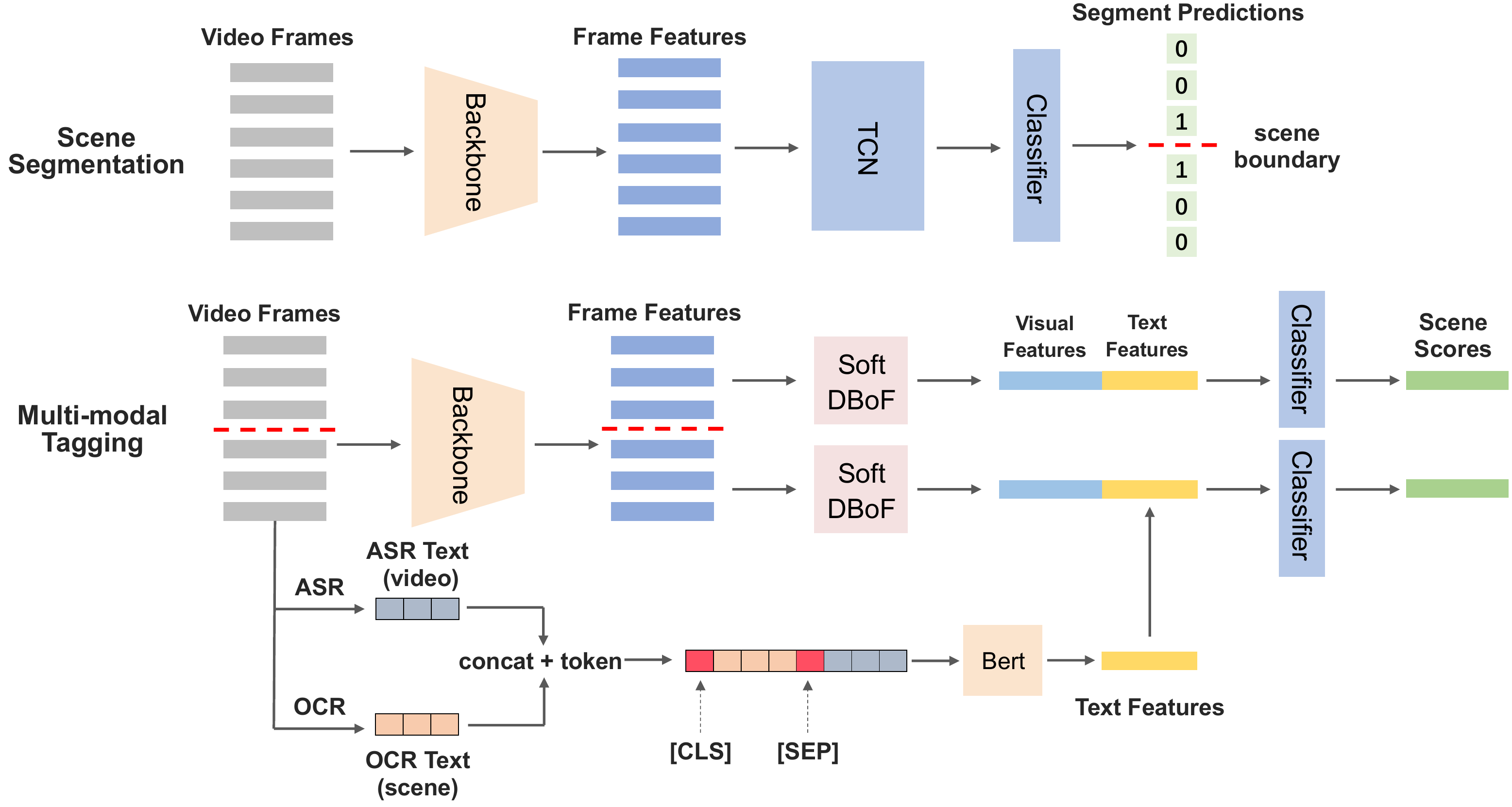}
\caption{An overview of our framework, which consists of scene segmentation and multi-modal tagging.}
\label{fig:framework}
\end{figure*}

\section{Approach}

Given a test video, our goal is to learn how to correctly segment video scenes and effectively make multi-label predictions for each scene, so as to analyze video ads at a fine-grained level.
We achieve this with a two-stage multi-modal analysis framework, which operates on different modalities, including audio dialogues, appearance clues, and textual information among key frames.
Figure~\ref{fig:framework} presents an overview of the framework. Below, we first introduce the preliminaries of our work, and then elaborate on how we solve the video ads content structuring problem in two stages, which are scene segmentation and multi-modal tagging.

\subsection{Preliminaries} \label{sec:pre}
Given a training set with $N$ video clips, where each video is associated with two types of modalities, \ie, audio and appearance.  Thus, we can represent a video with $T$ video snippets as:

\begin{align}
  \bm{V} = & [\bm{V}_1, \bm{V}_2, \bm{V}_t, \cdots, \bm{V}_T] \\
  \text{where} \, & \bm{V}_t =   \{\bm{V}^{A}_t, \bm{V}^{I}_t\}.
\end{align}

Here, $\bm{V}_t$ contains $\bm{V}^{A}_t and \bm{V}^{I}_t $ denote the audio and image modality. Below we introduce how to extract useful information from the two modalities.

\vspace{1.5mm}
\noindent \textbf{Audio Modality.}  Audio signals often contain important contextual information and are helpful to improve the accuracy of video classification. Because the audio information in video ads mainly exists in the form of dialogues, we use ASR to convert the original audio signals into texts, and then perform recognition based on recognized text from dialogues. However, extracting local ASR information in a short time window is challenging as short video clips tend to contain discontinuous speech and confusing context. 
While global ASR outputs for the entire video are provided by the organizers, they lack temporal annotations at the scene-level and it is challenging to perform grounding to associate texts with scenes. Since we need ASR clues for fine-grained understanding, we will introduce a global-local attention mechanism to mitigate this issue. Formally, we define the mapping from audio signals of a whole video to a speech record as $f_{\theta_A}:{V}^{A} \mapsto \bm{x}^{A}$, where $f_{\theta_A}$ denotes the weights of the audio speech recognition network and $\bm{x}^{A}$ is the speech.

\vspace{1.5mm}
\noindent \textbf{Image Modality. } We use a 2D CNN model to capture appearance information from RGB frames.
Compared to audio clues which are usually noisy, the appearance network provides decent visual information with a moderate computational cost and it suffices in most cases. To this end, we respectively use ResNet101, EfficientNet B4 and EfficientNet B5~\cite{tan2019efficientnet} backbone to extract appearance clues. 
Also, we adopt a 3D CNN model to capture motion clues that depict how objects move among stacked frames. We instantiate the motion network with a SlowFast network~\cite{feichtenhofer2019slowfast}. 
Formally, we define the mapping from a stack of frames or an RGB frame to a feature vector as $f_{\theta_I}: {V}^{I}_t \mapsto \bm{x}^{I}_t$, where $f_{\theta_I}$ is the weights of the network (ResNet101, EfficentNet B4, EfficientNet B5 or SlowFast) and $\bm{x}^{I}_t$ represents visual features. 

In addition, the 2D CNN models are pretrained on ImageNet and the 3D CNN model are pretrained on Kinetics-400, and then all the models are finetuned on the ads dataset. More details can be found in section~\ref{sec:impd}.

\vspace{1.5mm}
\noindent \textbf{Textual Information. } Optical characters in videos, such as subtitles and signboards, often provide rich information to improve video classification accuracy. We apply OC-OCR~\cite{Oc-OCR} to capture optical character information for each frame. 

Formally, we define the mapping from a frame to a set of character sequences as  $f_{\theta_T}: {V}^{I} \mapsto \bm{X}^{T}$, where $f_{\theta_T}$ is the weights of the OCR model, ${V}^{I}$ is a frame in video and $\bm{X}^{T}=\{\bm{x}^{T}_1, \bm{x}^{T}_2, ..., \bm{x}^{T}_N  \}$ is a set of recognition character sequence in the frame. Besides, optical character information between adjacent frames is often redundant. To overcome this problem, we propose a post-processing method to remove the same text in adjacent frames.

\subsection{Temporal Convolution Networks for Scene Segmentation}

\begin{figure}[h!]
\centering
\includegraphics[width=0.9\linewidth]{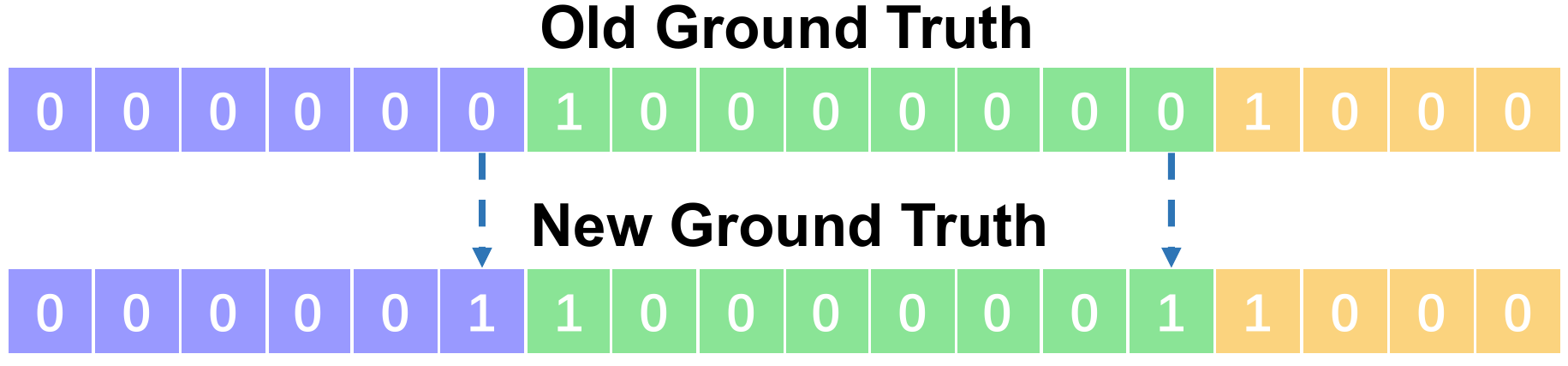}
\caption{We modify the original GT by setting both the beginning and the end of scenes as transition points.}
\label{fig:transGT}
\end{figure}

We build upon Temporal Convolution Network (TCN) due to its effectiveness for action segmentation. Briefly, we firstly stack two TCN blocks as the shallow layers of the network, and each TCN block consists of four 1-D dilated convolutional layers, whose dilated sizes are 1, 2, 4, 8 respectively. Then, intuitively, we use a normal 1-d convolution in the last layer of the network, to avoid confusing boundaries. We also use dropout to prevent overfitting. We consider the scene segmentation problem as a binary classification problem at each time step. Therefore, after we obtain the new representation of each time step by multiple TCNs, we will classify it through fully connected layers. Since ground-truth provided by the organizers only contain one transition point for each scene, we argue that such annotations will lead to biased training sets. We modify the original ground-truth labels by setting both the beginning and the end of scenes as transition points (See Fig.~\ref{fig:transGT} for an illustration).
Then we optimize the TCN network using a standard binary cross-entropy loss.  We further use focal loss~\cite{lin2017focal} for improved performance.

\subsection{Multi-Modal Tagging for Scenes}

Video ads are multi-modal in nature. In 2021 TAAC, we use image and textual information for multi-label video classification. For visual information, we introduce a  NeXt-SoftDBoF module for temporal aggregation. Compared with NeXtVLAD~\cite{lin2018nextvlad}, NeXt-SoftDBoF has fewer parameters and is less prone to overfitting. For text sequences, we design a global-local attention mechanism to combine ASR and OCR outputs. 
Finally, we concatenate the logits values from two modalities for improved performance.

\subsubsection{NeXt-SoftDBoF} \,

We build upon NeXtVLAD to construct our NeXt-SoftDBoF Aggregation network, which can be regarded as a variant of Soft-BoF(soft bag of features)\cite{miech2017learnable}. The core idea is to expand the original $K$ cluster centers to $G*K$ and guarantee the final word frequency statistics dimension is still $K$ with a multi-head operation. Specifically, the input features firstly complete the soft-word frequency statistics at the $K$ centers in each group, and then an attention module will assign different weights to the $G$ groups to achieve weighted summation. 

We firstly expand the input vector $x_i$ as $\dot{x_i}$ with a dimension of $\lambda \, N$ via a linear fully-connected layer, where $\lambda$ is the expansion multiple and $N$ is the dimension of $x_i$. Then NeXt-SoftDBoF is formalized as follows, 

\begin{align}
    y_{k} &= \sum_{\substack{ i\in\{1,...,T\} \\ g\in\{1,...,G\}}}  \alpha_g(\dot{x}_i)\alpha_{gk}(\dot{x}_i)\\
        \alpha_g({\dot{x}_i})&=\sigma(\hat{w}_g^T \dot{x}_i + \hat{b}_g)\\
    \alpha_{gk}(\dot{x}_i)&=\frac{e^{ w_{gk}^{T}\,\dot{x}_{i} + b_{gk}}}{\sum_{s=1}^{K} e^{ w_{gs}^{T}\,\dot{x}_{i} + b_gs}}
\end{align}

where $\alpha_g(\dot{x}_i)$ represents the attention weights of the g-th group and $\alpha_{gk}(\dot{x}_i) $ represents the "frequency" of the i-th feature belonging to the k-th cluster center in the g-th group. Finally, we get the result of aggregation as $\bm{y} = [y1, y2, ..., y_k]$. 

\begin{table}[h!]
\centering
  \caption{Comparisons between OCR and ASR. }
  \vspace{-0.1in}
  \label{tab:textCom}
  \begin{tabular}{c|cc}
    \toprule 
       text type& pros& cons \\
    \midrule
       OCR& has temporal locality & Noisy\\
       ASR& Concise and accurate & Global  \\
    \bottomrule
  \end{tabular}
\end{table}

\begin{table*}[h!]
\centering
  \caption{Performance. SF: SlowFast; E4: EfficientNet-B4; E5: EfficientNet-B5; R: ResNet101. ``n* '' indicates models produced by different seeds. If GL-Attention is \XSolidBrush, only OCR features will be used, else, global(ASR)-local(OCR) attention will be added.  }

    \vspace{-0.1in}
  \label{tab:main}
  \begin{tabular}{c||ccc|ccc|c||c}
    \toprule
    \multirow{2}{*}{Exp Name} & \multicolumn{3}{c|}{SceneSeg} & \multicolumn{3}{c|}{Multi-Modal Tagging} &  \multirow{2}{*}{threshold} & \multirow{2}{*}{F1*mAP@20} \\
    
    \cmidrule(lr){2-4}\cmidrule(lr){5-7}
    
    & Backbone & Finetune & Focal Loss & Backbone & Finetune& GL-Atttention & \\
    
    \midrule 
    Baseline & R & \XSolidBrush & \XSolidBrush & R &\XSolidBrush&\XSolidBrush& 0.5 &0.1565  \\
    \midrule 
    I & R & \XSolidBrush & \XSolidBrush & R &\Checkmark&\XSolidBrush& \multirow{7}{*}{0.5} &0.1894  \\
    II & R & \Checkmark & \XSolidBrush & R &\Checkmark&\XSolidBrush&&0.1964  \\
    III & R & \Checkmark & \XSolidBrush & SF &\Checkmark&\XSolidBrush&&0.2051  \\
    IV & R & \Checkmark & \XSolidBrush & R+SF &\Checkmark&\XSolidBrush&&0.2122  \\
    V & R & \Checkmark & \Checkmark & R+SF &\Checkmark&\XSolidBrush&&0.2141  \\
    VI & R & \Checkmark & \XSolidBrush & R+SF &\Checkmark&\Checkmark&&0.2230  \\
    VII & R+E4 & \Checkmark & \XSolidBrush & R+SF+E4 &\Checkmark&\Checkmark&&0.2327  \\
    \midrule
    VIII & 3*(R+E4+E5) & \Checkmark & \XSolidBrush & 1*R+2*E4+3*SF &\Checkmark&\Checkmark&0.5&0.2389  \\
    IX & 3*(R+E4+E5) & \Checkmark & \XSolidBrush & 1*R+2*E4+3*SF &\Checkmark&\Checkmark&0.45&0.2419  \\
    X & 4*(R+E4+E5) & \Checkmark & \XSolidBrush & 3*R+3*E4+6*SF &\Checkmark&\Checkmark&0.45&0.2438  \\
    XI & 4*(R+E4+E5) & \Checkmark & \XSolidBrush & 3*R+3*E4+6*SF &\Checkmark&\Checkmark&0.4&\textbf{0.2455}  \\
    XII & 4*(R+E4+E5) & \Checkmark & \XSolidBrush & 3*R+3*E4+6*SF &\Checkmark&\Checkmark&0.35&\textbf{0.2470}  \\
    \bottomrule
  \end{tabular}
\end{table*}

\subsubsection{Global-Local Attention} 
Texts in ads provide rich information. As mentioned in Section~\ref{sec:pre}, we have obtained two kinds of textual information: OCR in each frame and ASR for the entire video. The characters of OCR and ASR are summarized in Table~\ref{tab:textCom}. Texts from OCR are usually noisy or redundant. for example, in the ads of math homework guidance, OCR may recognize a bunch of math formulas, resulting in very noisy OCR outputs. 
ASR mainly records the dialogue of the characters, which is always relevant to the video content and is more concise and accurate than the OCR text. However, ASR text lacks temporal locality.

In order to solve these problems, we introduce a global-local attention mechanism, hoping that the global ASR and local OCR information can pay attention to each other and learn from each other. 
We concatenate the OCR text and the ASR text directly and use the <SEP> token to separate them.
More specifically, we follow the form of ``<CLS>[OCR text]<SEP>[ASR text]'' to construct the input text for the BERT model~\cite{devlin2018bert}.

For each scene, our model predicts the confidence scores of $C$ classes, where $C=82$ in TAAC 2021. Since scene classification is a multi-label task in TAAC 2021, we calculate BCE loss for each class, then sum them up to get the final tagging loss.

\section{Experiments}
\subsection{Experimental Setup.}
\subsubsection{Dataset.} 
The 2021 TAAC video ads dataset contains 5,000 training videos and 5000 testing videos, which are collected from online video advertisements. For each video, the boundary and the labels of each scene are provided as ground-truth annotations. To evaluate our method, we randomly sample 10\% videos from the training set as the local validation set and utilize the other data for training.

\subsubsection{Evaluation.}
$F1@0.5s$ is used for the evaluation of scene segmentation. Specifically, the predicted boundary of a scene segment is true positive if the gap between the ground-truth and the prediction is less than $0.5s$; otherwise, it is a false positive. Each ground truth only matches a prediction once. 

The predictions of scene tagging are evaluated by the mean Average Precision (mAP) with overlapping thresholds between 0.5 and 0.95 (stride 0.05), denoted by $average\ mAP@IoU=0.5:0.05:0.95$. The overlapping threshold is determined based on the temporal intersection over union (tIoU) ratio between predicted segments and ground-truth. For the evaluation of both the scene segmentation and the scene tagging, the final results are computed by the multiplication of F1-score and average mAP.

\subsubsection{Implementation details.} \label{sec:impd}

We train our models on Tesla V100 GPUs using PyTorch 1.7. We use the Adam optimizer~\cite{DBLP:journals/corr/KingmaB14} and the exponential moving average (EMA) with the weight of 0.9 for all training stages. When training models for scene segmentation, the learning rate is set as $1e^{-4}$ with a batch size of 64. When training models for scene classification, the learning rate is set to $1e^{-4}$ for the classifier and $1e^{-5}$ for other modules, and the batch size is 32. 

\subsection{Main Results and Discussion} 
The comparison results are summarized in Table~\ref{tab:main}.

\textbf{Baseline :} We use ResNet101 pretrained on ImageNet to capture frame features and it offers a score of 0.1565.

\textbf{Exp I :} We fine-tune the ResNet101 on TAAC 2021 scene classification and use its features for multi-label scene classification. The score is 0.1894, which achieves a significant increase of 0.0329 compared with the baseline. These results validate our hypothesis that fine-tuning model is beneficial to improve feature quality and scene classification accuracy.

\textbf{Exp II :} We utilize the fine-tuned ResNet101 features on the scene segmentation task, and also achieve higher performance. It indicates that fine-tuned features include better advertising semantics so that it can make better scene segmentation.

\textbf{Exp III :} In this experiment, we use fine-tuned SlowFast features for the scene classification task. It achieves a 0.087 score increase compared with fine-tuned ResNet101 features. Unlike 2D models, SlowFast captures more motion information. We have also tried 3D features, such as SlowFast and I3d, for scene segmentation. However, it doesn't work on our validation set. We assume that 3D models may not be suitable for scene segmentation since clip features are not temporally aligned.

\textbf{Exp IV :} We use ResNet101 features and SlowFast features for ensembling, which offers an 0.2122 score. It proves the effectiveness of model ensembling.

\textbf{Exp V:} We use focal loss when fine-tuning ResNet101 on the scene segmentation task. It only achieves a slight improvement. Limited by the number of competition submissions, we don't use it in our final solution.

\textbf{Exp VI:} In this experiment, we use ASR text and apply GL-Attention between ASR text and OCR text. It achieves a 0.0108 score increase, which proves ASR is effective for scene understanding. 

\textbf{Exp VII :} Based on Exp VI, we use EfficientNet B4 for both two tasks, which achieves almost 1 $\%$ improvement. It indicates that ensemble is also useful for scene segmentation. Besides, ensembling more features can improve performance.

\textbf{Exp VIII - XII:} To boost the performance further, we set different seeds to diversify model initialization and the train/val split. Meanwhile, we study the influence of the threshold on the results and finally achieve a score of 0.2470.

\section{Conclusion}
In this paper, we introduce a multimodal framework for video ads structuring understanding, which can effectively segment and predict video scenes. The system leverages global ASR information, local OCR information and visual clues for improved performance. The system ranked fourth in the 2021 TAAC leaderboard, which proves its effectiveness.

\bibliographystyle{unsrt}
\bibliography{sample-base}

\end{document}